\documentclass[11pt,a4paper]{article}
\usepackage[hyperref]{acl2018}
\usepackage{times}
\usepackage{latexsym}
\usepackage{tikz}
\usetikzlibrary{positioning,shapes,shadows,arrows}

\usetikzlibrary{trees}
\usepackage{amsmath}
\usepackage{comment}
\usepackage{enumitem}
\usepackage{tabularx}
\usepackage{array}

\usepackage{booktabs}
\newcommand{\tabitem}{~~\llap{\textendash}~~}

\tikzstyle{every node}=[draw=black,thick,anchor=west,rounded corners,drop shadow,fill=white]
\tikzstyle{selected}=[draw=red,fill=red!30]
\tikzstyle{optional}=[dashed,fill=gray!50]

\newenvironment{tight_enumerate}{
\begin{enumerate}[leftmargin=4.0mm]
  \setlength{\itemsep}{0pt}
  \setlength{\parskip}{0pt}
}{\end{enumerate}}

\usepackage{url}

\aclfinalcopy 

\title{A Corpus with Multi-Level Annotations of Patients, Interventions and Outcomes to Support Language Processing for Medical Literature}

\author{Benjamin Nye \\
  Northeastern University \\
  {\small\tt nye.b@husky.neu.edu}  \\\And 
  Junyi Jessy Li \\ 
  UT Austin \\ 
  {\small\tt jessy@austin.utexas.edu} \\\And
  Roma Patel \\
  Rutgers University\\
  {\small\tt romapatel996@gmail.com} \\\AND
  Yinfei Yang\thanks{* now at Google Inc.} \\
  \emph{No affiliation} \\
  {\small\tt yangyin7@gmail.com} \\ \\\And 
  Iain J. Marshall \\
  King's College London \\
  {\small\tt iain.marshall@kcl.ac.uk} \\\And
  Ani Nenkova \\
  UPenn \\
  {\small\tt nenkova@seas.upenn.edu} \\\AND
  Byron C. Wallace \\
  Northeastern University \\
  {\small\tt b.wallace@northeastern.edu}}
  
\date{}

\begin{document}
\maketitle
\begin{abstract}

We present a corpus of 5,000 richly annotated abstracts of medical articles describing 
clinical randomized controlled trials. Annotations include demarcations of text spans that describe the Patient population enrolled, the Interventions studied and to what they were Compared, and the Outcomes measured (the `PICO' elements). These spans are further annotated at a more granular level, e.g., individual interventions within them are marked and mapped onto a structured medical vocabulary. We acquired annotations from a diverse set of workers with varying levels of expertise and cost. We describe our data collection process and the corpus itself in detail. We then outline a set of challenging NLP tasks that would aid searching of the medical literature and the practice of evidence-based medicine. 
\end{abstract}

\section{Introduction}
\label{section:intro}
 
In 2015 alone, about 100 manuscripts describing randomized controlled trials (RCTs) for medical interventions were published \emph{every day}. It is thus practically impossible for physicians to know which is the best medical intervention for a given patient group and condition  \cite{borah2017analysis,fraser2010impossibility,bastian2010seventy}.
This inability to easily search and organize the published literature impedes the aims of \emph{evidence based medicine} (EBM), which aspires to inform patient care using the totality of 
relevant evidence. 
Computational methods could expedite biomedical evidence synthesis \cite{tsafnat2013automation,wallace2013modernizing} and natural language processing (NLP) in particular can play a key role in the task. 


Prior work has explored the use of NLP methods to automate biomedical evidence extraction and synthesis \cite{boudin2010positional,marshall:2017:ACL,ferracane2016leveraging,verbeke2012statistical}.\footnote{There is even, perhaps inevitably, a systematic review of such approaches \cite{jonnalagadda2015automating}.} But the area has attracted less attention than it might from the NLP community, due primarily to a dearth of publicly available, annotated corpora with which to train and evaluate models. 

Here we address this gap by introducing \emph{EBM-NLP}, a new corpus to power NLP models in support of EBM. The corpus, accompanying documentation, baseline model implementations for the proposed tasks, and all code are publicly available.\footnote{\url{http://www.ccs.neu.edu/home/bennye/EBM-NLP}} EBM-NLP comprises $\sim$5,000 medical abstracts describing clinical trials, multiply annotated in detail with respect to characteristics of the underlying trial Populations (e.g., \emph{diabetics}), Interventions (\emph{insulin}), Comparators (\emph{placebo}) and Outcomes (\emph{blood glucose levels}). Collectively, these key informational pieces are referred to as PICO elements; they form the basis for well-formed clinical questions \cite{huang2006evaluation}. 

We adopt a hybrid crowdsourced labeling strategy using heterogeneous annotators with varying expertise and cost, from laypersons to MDs. Annotators were first tasked with marking text spans that described the respective PICO elements. Identified spans were subsequently annotated in greater detail: this entailed finer-grained labeling of PICO elements and mapping these onto a normalized vocabulary, and indicating redundancy in the mentions of PICO elements. 

In addition, we outline several NLP tasks that would directly support the practice of EBM and that may be explored using the introduced resource. We present baseline models and associated results for these tasks.

\section{Related Work}
\label{section:related-work}

We briefly review two lines of research relevant to the current effort: work on NLP to facilitate EBM, and research in crowdsourcing for NLP. 

\subsection{NLP for EBM}
Prior work on NLP for EBM has been limited by the availability of only small corpora, which have typically provided on the order of a couple hundred annotated abstracts or articles for very complex information extraction tasks. For example, the ExaCT system \cite{kiritchenko2010exact} applies rules to extract 21 aspects of the reported trial. It was developed and validated on a dataset of 182 marked full-text articles. The ACRES system \cite{summerscales2011automatic} produces summaries of several trial characteristic, and was 
trained on 263 annotated abstracts.
Hinting at more challenging tasks that can build upon foundational information extraction, Alamri and Stevenson \shortcite{alamri2015automatic} developed methods for detecting contradictory claims in biomedical papers. Their corpus of annotated claims contains 259 sentences \cite{alamri2016corpus}.



Larger corpora for EBM tasks have been derived using (noisy) automated annotation approaches. This approach has been used to build, e.g., datasets to facilitate work on Information Retrieval (IR) models for biomedical texts 
\cite{scells2017test,chung2009sentence,boudin2010positional}. Similar approaches have been used to `distantly supervise' annotation of full-text articles describing clinical trials \cite{wallace2016extracting}. In contrast to the corpora discussed above, these automatically derived datasets tend to be relatively large, but they include only shallow annotations.




Other work attempts to bypass basic extraction tasks and address more complex biomedical QA and (multi-document) summarization problems to support EBM \cite{demner2007answering,molla2011development,abacha2015means}. Such systems would directly benefit from more accurate extraction of the types 
codified in the corpus we present here.


\subsection{Crowdsourcing}

Crowdsourcing, which we here define operationally as the use of distributed lay annotators, has shown encouraging results in NLP \cite{novotney2010cheap,sabou2012crowdsourcing}. Such annotations are typically imperfect, but methods that aggregate redundant annotations can mitigate this problem \cite{dalvi2013aggregating,hovy2014experiments,nguyen2017aggregating}.

Medical articles contain relatively technical content, which intuitively may be difficult for persons without domain expertise to annotate. However, recent promising preliminary work has found that crowdsourced approaches can yield surprisingly high-quality annotations in the domain of EBM specifically \cite{mortensen2017exploration,thomas2017living,wallace2017identifying}.


\section{Data Collection}
\label{section:collection}

\begin{figure*}
\centering
\includegraphics[width=0.8\linewidth]{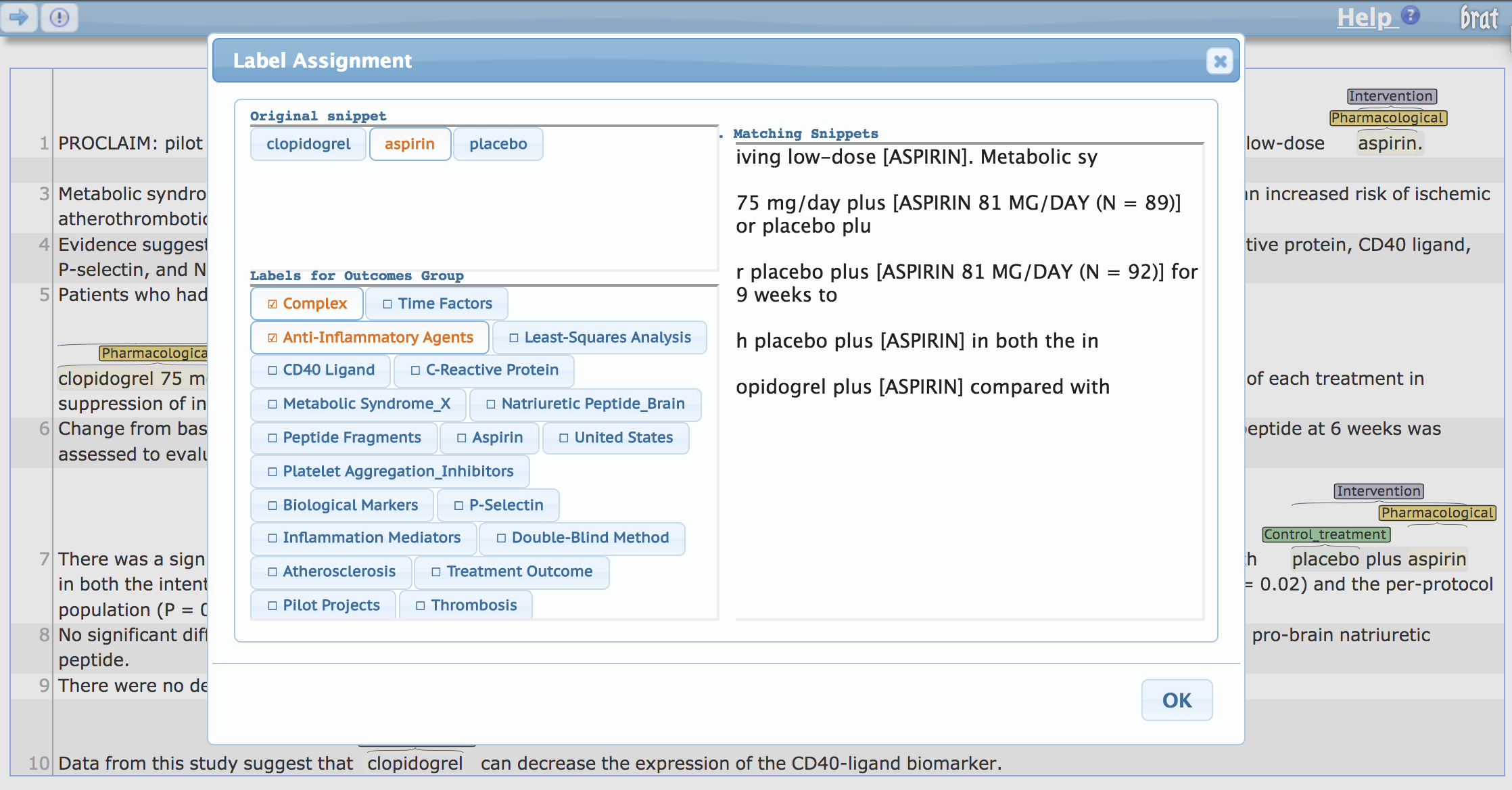}
\caption{Annotation interface for assigning MeSH terms to snippets.}
\end{figure*}


PubMed provides access to the MEDLINE database\footnote{\url{https://www.nlm.nih.gov/bsd/pmresources.html}} which indexes titles, abstracts and meta-data for articles from selected medical journals dating back to the 1970s. MEDLINE indexes over 24 million abstracts; the majority of these have been manually assigned metadata which we used to retrieved a set of 5,000 articles describing RCTs with an emphasis on cardiovascular diseases, cancer, and autism. These particular topics were selected to cover a range of common conditions.



We decomposed the annotation process into two steps, performed in sequence. First, we acquired labels demarcating spans in the text describing the clinically salient abstract elements mentioned above: the trial Population, the Interventions and Comparators studied, and the Outcomes measured. We collapse Interventions and Comparators into a single category (I).
In the second annotation step, we tasked workers with providing more granular (sub-span) annotations on these spans. 

For each PIO element, all abstracts were annotated with the following four types of information.
\begin{enumerate}
\item \textbf{Spans} exhaustive marking of text spans containing information relevant to the respective PIO categories (Stage 1 annotation).
\item \textbf{Hierarchical labels} assignment of more specific labels to subsequences comprising the marked relevant spans (Stage 2 annotation).
\item \textbf{Repetition} grouping of labeled tokens to indicate repeated occurrences of the same information (Stage 2 annotation).
\item \textbf{MeSH terms} assignment of the metadata MeSH terms associated with the abstract to labeled subsequences (Stage 2 annotation).\footnote{MeSH is a controlled, structured medical vocabulary maintained by the National Library of Medicine.}   
\end{enumerate}

We collected annotations for each P, I and O element individually to avoid the cognitive load imposed by switching between label sets, and to reduce the amount of instruction required to begin the task. All annotation was performed using a modified version of the Brat Rapid Annotation Tool (BRAT) \citep{stenetorp2012brat}. We include all annotation instructions provided to workers for all tasks in the Appendix.


\subsection{Non-Expert (Layperson) Workers}

For large scale crowdsourcing via recruitment of layperson annotators, we used Amazon Mechanical Turk (AMT). All workers were required to have an overall job approval rate of at least 90\%. Each job presented to the workers required the annotation of three randomly selected abstracts from our pool of documents. As we received initial results, we blocked workers who were clearly not following instructions, and we actively recruited the best workers to continue working on our task at a higher pay rate. 

We began by collecting the least technical annotations, moving on to more difficult tasks only after restricting our pool of workers to those with a demonstrated aptitude for the jobs. We obtained annotations from $\geq 3$ different workers for each of the 5,000 abstracts to enable robust inference of reliable labels from noisy data. After performing filtering passes to remove non-RCT documents or those missing relevant data for the second annotation task, we are left with between 4,000 and 5,000 sets of annotations for each PIO element after the second phase of annotation. 

\subsection{Expert Workers}

To supplement our larger-scale data collection via AMT, we collected annotations for 200 abstracts for each PIO element from workers with advanced medical training. The idea is for these to serve as reference annotations, i.e., a test set with which to evaluate developed NLP systems. We plan to enlarge this test set in the near future, at which point we will update the website accordingly. 

For the initial span labeling task, two medical students from the University of Pennsylvania and Drexel University provided the reference labels. 
In addition, for both stages of annotation and for the detailed subspan annotation in Stage 2, we hired three medical professionals via Upwork,\footnote{\url{http://www.upwork.com}} an online platform for hiring skilled freelancers. 
After reviewing several dozen suggested profiles, we selected three workers that had the following characteristics: Advanced medical training (the majority of hired workers were Medical Doctors, the one exception being a fourth-year medical student); Strong technical reading and writing skills; And an interest in medical research. 
In addition to providing high-quality annotations, individuals hired via Upwork also provided feedback regarding the instructions to help make the task as clear as possible for the AMT workers.

\section{The Corpus}
\label{section:the-corpus}

We now present corpus details, paying special attention to worker performance and agreement. We discuss and present statistics for acquired annotations on spans, tokens, repetition and MeSH terms in Sections \ref{section:corpus-spans}, \ref{section:corpus-tokens}, \ref{section:corpus-coref}, and \ref{section:corpus-mesh}, respectively.

\begin{table*} 
\centering
\small
\begin{tabularx}{\linewidth}{ m{0.1cm} m{2.6cm} l X }
\noalign{\vskip 1mm}  
\multicolumn{4}{l}{\textbf{P} \emph{Fourteen children (12 infantile autism full syndrome present, 2 atypical pervasive developmental disorder) between 5 and}} \\
\multicolumn{4}{l}{\emph{13 years of age}} \\
\noalign{\vskip 1mm}  

& Text & Label & MeSH terms \\
\cline{2-4}
\noalign{\vskip 1mm}  
& \tabitem Fourteen & \textsc{Sample Size (full)}    & \\
& \tabitem children & \textsc{Age (young)}              & \\
& \tabitem 12       & \textsc{Sample Size (partial)} & \\
& \tabitem autism   & \textsc{Condition (disease)}   & Autistic Disorder, Child Development Disorders Pervasive \\
& \tabitem 2        & \textsc{Sample Size (partial)} & \\
& \tabitem 5 and 13 & \textsc{Age (young)}             & \\

\noalign{\vskip 2mm}  
\multicolumn{4}{l}{\textbf{I} \emph{20 mg Org 2766 (synthetic analog of ACTH 4-9)/day during 4 weeks, or placebo in a randomly assigned sequence.}} \\
\noalign{\vskip 1mm}  

& Text & Label & MeSH terms \\
\cline{2-4}
\noalign{\vskip 1mm}  
& \tabitem 20 mg Org 2766    & \textsc{Pharmacological} & Adrenocorticotropic Hormone, Double-Blind Method, Child Development Disorders Pervasive\\
& \tabitem placebo           & \textsc{Control}         & Double-Blind Method\\

\noalign{\vskip 2mm}  
\multicolumn{4}{l}{\textbf{O} \emph{Drug effects and Aberrant Behavior Checklist ratings}} \\
\noalign{\vskip 1mm}  

& Text & Label & MeSH terms \\
\cline{2-4}
\noalign{\vskip 1mm}  
& \tabitem Drug effects                        & \textsc{Quality of Intervention} & \\
& \tabitem Aberrant Behavior Checklist ratings & \textsc{Mental (Behavior)}          & Attention,  Stereotyped Behavior\\
\hline
\end{tabularx}
\caption{Partial example annotation for Participants, Interventions, and Outcomes. The full annotation includes multiple top-level spans for each PIO element as well as labels for repetition. }
\end{table*}


\subsection{Spans}
\label{section:corpus-spans}
For each P, I and O element, workers were asked to read the abstract and highlight all spans of text including any pertinent information. Annotations for 5,000 articles were collected from a total of 579 AMT workers across the three annotation types, and expert annotations were collected for 200 articles from two medical students. 

\begin{table}[h]
    \centering
    \small
    \begin{tabular}{ l c } 
         & Agreement \\
         \cline{2-2}
         Participants & 0.71 \\
         Interventions & 0.69 \\
         Outcomes & 0.62 \\
    \end{tabular}
    \caption{Cohen's $\kappa$ between medical students for the 200 reference documents.}
     \label{tab:span_agreement}
\end{table}

We first evaluate the quality of the annotations by calculating token-wise label agreement between the expert annotators; this is reported in Table \ref{tab:span_agreement}.
Due to the difficulty and technicality of the material, agreement between even well-trained domain experts is imperfect.
The effect is magnified by the unreliability of AMT workers, motivating our strategy of collecting several noisy annotations and aggregating over them to produce a single cleaner annotation.
We tested three different aggregation strategies: a simple majority vote, the Dawid-Skene model \cite{dawid1979maximum} which estimates worker reliability, 
and HMMCrowd, a recent extension to Dawid-Skene that includes a HMM component, thus explicitly leveraging the sequential structure of contiguous spans of words \citep{nguyen2017aggregating}.
\begin{table}
    \centering
    \small
    \begin{tabular}{  l c c c  }
        \hline
        \textbf{Participants} & Precision & Recall & F-1 \\
        \cline{2-4}
        Majority Vote   & 0.903  & 0.507  & 0.604  \\
        Dawid Skene     & 0.840  & 0.641  & 0.686  \\
        HMMCrowd        & 0.719  & 0.761  & 0.698  \\
        \hline
        \textbf{Interventions} & Precision & Recall & F-1 \\
        \cline{2-4}
        Majority Vote   & 0.843  & 0.432  & 0.519  \\
        Dawid Skene     & 0.755  & 0.623  & 0.650  \\
        HMMCrowd        & 0.644  & 0.800  & 0.683  \\
        \hline
        \textbf{Outcomes} & Precision & Recall & F-1 \\
        \cline{2-4}
        Majority Vote   & 0.711  & 0.577  & 0.623  \\
        Dawid Skene     & 0.652  & 0.648  & 0.629  \\
        HMMCrowd        & 0.498  & 0.807  & 0.593  \\
    \end{tabular}
    \caption{Precision, recall and F-1 for aggregated AMT spans evaluated against the union of expert span labels, for all three P, I, and O elements.}
    \label{tab:basic_exp}
\end{table}

For each aggregation strategy, we compute the token-wise precision and recall of the output labels against the unioned expert labels.
As shown in Table~\ref{tab:basic_exp}, the HMMCrowd model afforded modest improvement in F-1 scores over the standard Dawid-Skene model, and was thus used to generate the inputs for the second annotation phase.

The limited overlap in the document subsets annotated by any given pair of workers, and wide variation in the number of annotations per worker make interpretation of standard agreement statistics tricky. We quantify the centrality of the AMT span annotations by calculating token-wise precision and recall for each annotation against the aggregated version of the labels (Table \ref{tab:amt_span_stats}).


\begin{table}[h]
    \centering
    \small
    \begin{tabular}{ l c c c } 
        \textbf{} & Precision & Recall & F-1\\
        \cline{2-4}
        Participants  & 0.34 & 0.29 & 0.30 \\
        Interventions & 0.20 & 0.16 & 0.18 \\ 
        Outcomes      & 0.11 & 0.10 & 0.10 \\ 
    \end{tabular}
    \caption{Token-wise statistics for individual AMT annotations evaluated against the aggregated versions.}
   	\label{tab:amt_span_stats}
\end{table}

When comparing the average precision and recall for individual crowdworkers against the aggregated labels in Table~\ref{tab:amt_span_stats}, scores are poor showing very low agreement between the workers.
Despite this, the aggregated labels compare favorably against the expert labels. This further supports the intuition that it is feasible to collect multiple low-quality annotations for a document and synthesize them to extract the signal from the noise.

On the dataset website, we provide a variant of the corpus that includes all individual worker span annotations (e.g., for researchers interested in crowd annotation aggregated methods), and also a version with pre-aggregated annotations for convenience.


%



\begin{table}[h]
    \centering
    \small
    \begin{tabular}{ l c c } 
        & \multicolumn{2}{l}{\textbf{Span frequency}} \\
        \textbf{} & AMT & Experts \\
        \cline{2-3}
        Participants  & 34.5 & 21.4 \\
        Interventions & 26.5 & 14.3 \\ 
        Outcomes      & 33.0 & 26.9 \\ 
    \end{tabular}
    \caption{Average per-document frequency of different token labels.}
   	\label{tab:span_label_freq}
\end{table}

\subsection{Hierarchical Labels}
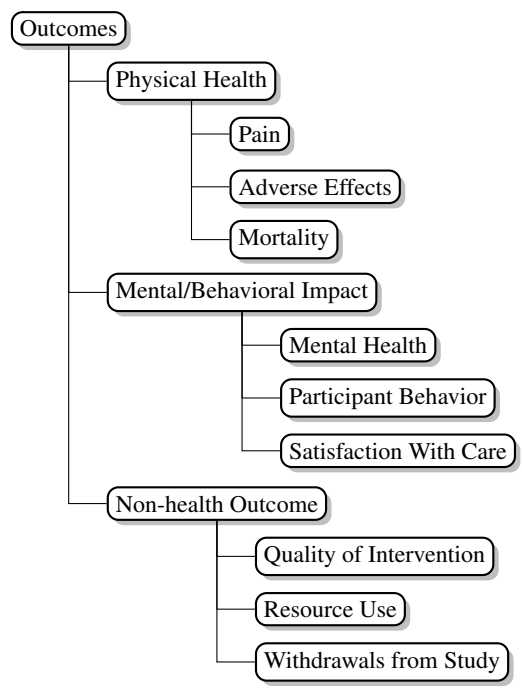
\begin{figure}
\centering
\small
\label{o_labels}
\begin{tikzpicture}[%
  grow via three points={one child at (0.5,-0.7) and
  two children at (0.5,-0.7) and (0.5,-1.4)},
  edge from parent path={(\tikzparentnode.south) |- (\tikzchildnode.west)}]
  \node {Outcomes}
    child { node {Physical Health}
    	child { node{Pain}}
    	child { node{Adverse Effects}}
    	child { node{Mortality}}
    }		
    child [missing] {}
    child [missing] {}
    child [missing] {}
    child { node {Mental/Behavioral Impact}
      child { node{Mental Health}}
      child { node{Participant Behavior}}
      child { node{Satisfaction With Care}}
    }
    child [missing] {}
    child [missing] {}
    child [missing] {}
    child { node {Non-health Outcome}
      child { node {Quality of Intervention}}
      child { node {Resource Use}}
      child { node {Withdrawals from Study}}
    };
\end{tikzpicture}
\caption{Outcome task label hierarchy}
\label{fig:outcome-hierarchy}
\end{figure}

\label{section:corpus-tokens}
For each P, I, and O category we developed a hierarchy of labels intended to capture important sub categories within these.
Our labels are aligned to (and thus compatible with) the concepts codified by the Medical Subject Headings (MeSH) vocabulary of medical terms maintained by the National Library of Medicine (NLM).\footnote{\url{https://www.nlm.nih.gov/mesh/}} In consultation with domain experts, we selected subsets of MeSH terms for each PIO category that captured relatively precise information without being overwhelming. 
For illustration, we show the outcomes label hierarchy we used in Figure \ref{fig:outcome-hierarchy}.
We reproduce the label hierarchies used for all PIO categories in the Appendix. 

At this stage, workers were presented with abstracts in which relevant spans were highlighted, based on the annotations collected in the first annotation phase (and aggregated via the HMMCrowd model).
This two-step approach served dual purposes: (i) increasing the rate at which workers could complete tasks, and (ii) improving recall by directing workers to all areas in abstracts where they might find the structured information of interest. 
Our choice of a high recall aggregation strategy for the starting spans ensured that the large majority of relevant sections of the article were available as inputs to this task.

The three trained medical personnel hired via Upwork each annotated 200 documents and reported that spans sufficiently captured the target information.
These domain experts received feedback and additional training after labeling an initial round of documents, and all annotations were reviewed for compliance.
The average inter-annotator agreement is reported in Table~\ref{tab:semantic_agreement}.

\begin{table}[h]
    \centering
    \small
    \begin{tabular}{ l c } 
        \textbf{} & Agreement \\
        \cline{2-2}
        Participants  & 0.50 \\
        Interventions & 0.59 \\ 
        Outcomes      & 0.51 \\ 
    \end{tabular}
    \caption{Average pair-wise Cohen's $\kappa$ between three medical experts for the 200 reference documents.}
   	\label{tab:semantic_agreement}
\end{table}

With respect to crowdsourcing on AMT, the task for Participants was published first, allowing us to target higher quality workers for the more technical Interventions and Outcomes annotations.  
We retained labels from 118 workers for Participants, the top 67 of whom were invited to continue on to the following tasks.
Of these, 37 continued to contribute to the project.
Several workers provided $\geq$ 1,000 annotations and continued to work on the task over a period of several months.

To produce final per-token labels, we again turned to aggregation.
The subspans annotated in this second pass were by construction shorter than the starting spans, and (perhaps as a result) informal experiments revealed little benefit from HMMCrowd's sequential modeling aspect.
The introduction of many label types significantly increased the complexity of the task, resulting in both lower expert inter-annotator agreement (Table~\ref{tab:semantic_agreement} and decreased performance when comparing the crowdsourced labels against those of the experts (Table~\ref{tab:semantic_aggregation}.

\begin{table}[h]
    \centering
    \small
    \begin{tabular}{  l c c c  }
        \hline
        \noalign{\vskip 1mm}  
        \textbf{Participants} & Precision & Recall & F-1 \\
        \cline{2-4}
        \noalign{\vskip 1mm}  
        Majority Vote   & 0.46  & 0.58  & 0.51  \\
        Dawid Skene & 0.66 & 0.60 & 0.63 \\
        \hline
        \noalign{\vskip 1mm}  
        \textbf{Interventions} & Precision & Recall & F-1 \\
        \cline{2-4}
        \noalign{\vskip 1mm}  
        Majority Vote & 0.56 & 0.49 & 0.52 \\
        Dawid Skene & 0.56 & 0.52 & 0.54 \\
        \hline
        \noalign{\vskip 1mm}  
        \textbf{Outcomes} & Precision & Recall & F-1 \\
        \cline{2-4}
        \noalign{\vskip 1mm}  
        Majority Vote   & 0.73  & 0.69  & 0.71  \\
        Dawid Skene & 0.73 & 0.80 & 0.76  \\
    \end{tabular}
    \caption{Precision, recall, and F-1 for AMT labels against expert labels using different aggregation strategies.}
    \label{tab:semantic_aggregation}
\end{table}

\begin{figure}
\centering
\includegraphics[width=0.35\textwidth]{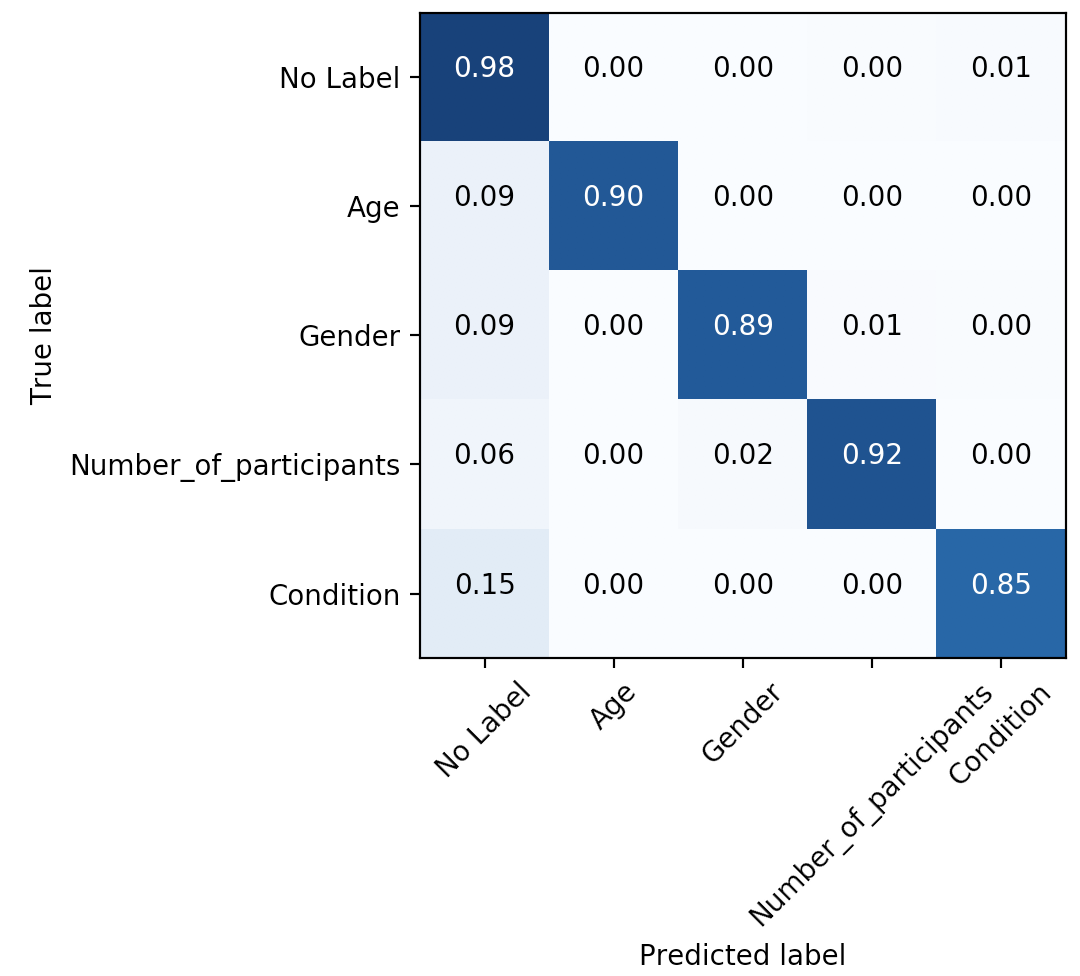}
\caption{Confusion matrix for token-level labels provided by experts.}
\label{fig:p_confusion}
\end{figure}

Most observed token-level disagreements (and errors, with respect to reference annotations) involve differences in the span lengths demarcated by individuals. 
For example, many abstracts contain an information-dense description of the patient population, focusing on their medical condition but also including information about their sex and/or age.
Workers would also sometimes fail to capture repeated mentions of the same information, producing Type 2 errors more frequently than Type 1.
This tendency can be seen in the overall token-level confusion matrix for AMT workers on the Participants task, shown in Figure~\ref{fig:p_confusion}. %

In a similar though more benign category of error, workers differed in the amount of context they included surrounding each subspan.
Although the instructions asked workers to highlight minimal subspans, there was variance in what workers considered relevant.

\begin{table}[h]
    \centering
    \small
    \begin{tabular}{ l c c c } 
        \textbf{} & Precision & Recall & F-1 \\
        \cline{2-4}
        Participants  & 0.39 & 0.71 & 0.50 \\
        Interventions & 0.59 & 0.60 & 0.60 \\ 
        Outcomes      & 0.70 & 0.68 & 0.69 \\ 
    \end{tabular}
    \caption{Statistics for individual AMT annotations evaluated against the aggregated versions, macro-averaged over different labels.}
   	\label{tab:amt_semantic_stats}
\end{table}

For the same reasons mentioned above (little pairwise overlap in annotations, high variance with respect to annotations per worker), quantifying agreement between AMT workers is again difficult using traditional measures. We thus again take as a measure of agreement the precision, recall, and F-1 of the individual annotations against the aggregated labels and present the results in Table~\ref{tab:amt_semantic_stats}.

\begin{table}[h]
    \centering
    \small
    \begin{tabular}{ l c c } 
        & \multicolumn{2}{l}{\textbf{Span frequency}} \\
        \cline{1-3}
        \textbf{Participants} & AMT & Experts\\
        \textsc{total} & 3.45 & 6.25 \\
        \cline{2-3}
        \noalign{\vskip 1mm}  
        Age & 0.49 & 0.66 \\
        Condition & 1.77 & 3.69 \\
        Gender & 0.36 & 0.34 \\
        Sample Size & 0.83 & 1.55 \\
        \hline
        \noalign{\vskip 1mm}  
        \textbf{Interventions} & AMT & Experts\\ 
        \textsc{total} & 6.11 & 9.31 \\
        \cline{2-3}
        \noalign{\vskip 1mm}  
        Behavioral & 0.22 & 0.37 \\
        Control & 0.83 & 0.94 \\
        Educational & 0.04 & 0.07 \\
        No Label & 0.00 & 0.00 \\
        Other & 0.23 & 1.12 \\
        Pharmacological & 3.37 & 5.19 \\
        Physical & 0.87 & 0.88 \\
        Psychological & 0.29 & 0.19 \\
        Surgical & 0.24 & 0.62 \\
        \hline
        \noalign{\vskip 1mm}  
        \textbf{Outcomes} & AMT & Experts\\
        \textsc{total} & 6.36 & 10.00 \\
        \cline{2-3}
        \noalign{\vskip 1mm}  
        Adverse effects & 0.45 & 0.66 \\
        Mental & 0.69 & 0.79 \\
        Mortality & 0.23 & 0.33 \\
        Other & 1.77 & 3.70 \\
        Pain & 0.18 & 0.27 \\
        Physical & 3.03 & 4.25 \\
    \end{tabular}
    \caption{Average per-document frequency of different label types.}
   	\label{tab:semantic_label_freq}
\end{table}

\subsection{Repetition}
\label{section:corpus-coref}

Medical abstracts often mention the same information in multiple places.
In particular, interventions and outcomes are typically described at the beginning of an abstract when introducing the purpose of the underlying study, and then again when discussing methods and results. It is important to be able to differentiate between novel and reiterated information, especially in cases such as complex interventions, distinct measured outcomes, or multi-armed trials. Merely identifying all occurrences of, for example, a pharmacological intervention leaves ambiguity as to how many distinct interventions were applied.

Workers identified repeated information as follows. After completing detailed labeling of abstract spans, they were asked to group together subspans that were instances of the same information (for example, redundant mentions of a particular drug evaluated as one of the interventions in the trial).
This process produces labels for repetition between short spans of tokens.
Due to the differences in the lengths of annotated subspans discussed in the preceding section, the labels are not naturally comparable between workers without directly modeling the entities contained in each subspan.
The labels assigned by workers produce repetition labels between sets of tokens but a more sophisticated notion of co-reference is required to identify which tokens correctly represent the entity contained in the span, and which tokens are superfluous noise.

As a proxy for formally enumerating these entities, we observe that a large majority of starting spans only contain a single target relevant to the subspan labeling task, and so identifying repetition between the starting spans is sufficient.
For example, consider the starting intervention span \emph{"underwent conventional total knee arthroplasty"}; there is only one intervention in the span but some annotators assigned the \textsc{surgical} label to all five tokens while others opted for only \emph{"total knee arthroplasty."}
By analyzing repetition at the level of the starting spans, we can compute agreement without concern for the confounds of slight misalignments or differences in length of the subspans.

Overall agreement between AMT workers for span-level repetition, measured by computing precision and recall against the majority vote for each pair of spans, is reported in Table~\ref{table:coref_acc}.

\begin{table}
    \centering
    \small
    \begin{tabular}{  l c c c  }
        & Precision & Recall & F-1 \\
        \cline{2-4}
        \noalign{\vskip 1mm}  
        Participants & 0.40 & 0.77 & 0.53 \\
        Interventions & 0.63 & 0.90 & 0.74 \\
        Outcomes & 0.47 & 0.73 & 0.57 \\
    \end{tabular}
    \caption{Comparison against the majority vote for span-level repetition labels.}
    \label{table:coref_acc}
\end{table}

\subsection{MeSH Terms}
\label{section:corpus-mesh}
The National Library of Medicine maintains an extensive hierarchical ontology of medical concepts called Medical Subject Headings (MeSH terms); this is part of the overarching Metathesaurus of the Unified Medical Language System (UMLS).
Personnel at the NLM manually assign citations (article titles, abstracts and meta-data) indexed in MEDLINE relevant MeSH terms. These terms have been used extensively to evaluate the content of articles, and are frequently used to facilitate document retrieval \cite{lu2009evaluation,lowe1994understanding}. 

In the case of randomized controlled trials, MeSH terms provide structured information regarding key aspects of the underlying studies, ranging from participant demographics to methodologies to co-morbidities. A drawback to these annotations, however, is that they are applied at the document (rather than snippet or token) level. To capture where MeSH terms are instantiated within a given abstract text, we provided a list of all terms associated with said article and instructed workers to select the subset of these that applied to each set of token labels that they annotated.

MeSH terms are domain specific and many require a medical background to understand, thus rendering this facet of the annotation process particularly difficult for untrained (lay) workers.
Perhaps surprisingly, several AMT workers voluntarily mentioned relevant background training; our pool of workers included (self-identified) nurses and other trained medical professionals. A few workers with such training stated this background as a reason for their interest in our tasks. 

\begin{figure}
\centering
\includegraphics[width=0.475\textwidth]{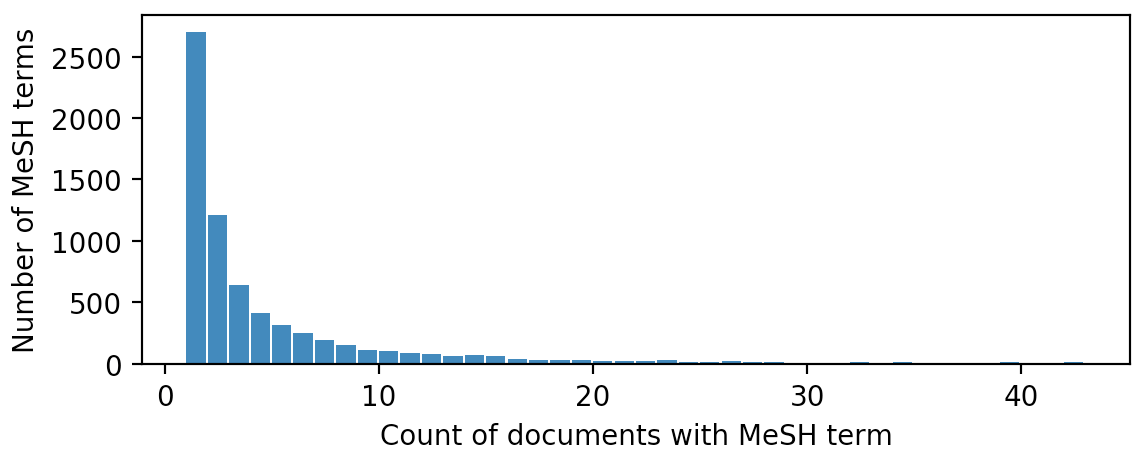}
\caption{Histogram of the number of documents containing each MeSH term.}
\label{fig:mesh_freq}
\end{figure}

The technical specificity of the more obscure MeSH terms is also exacerbated by their sparsity.
Of the 6,963 unique MeSH terms occurring in our set of abstracts, 87\% of them are only found in 10 documents or fewer and only 2.0\% occur in at least 1\% of the total documents.
The full distribution of document frequency for MeSH terms is show in Figure~\ref{fig:mesh_freq}.

To evaluate how often salient MeSH terms were instantiated in the text by annotators we consider only the 135 MeSH terms that occur in at least 1\% of abstracts (we list these in the supplementary material).
For each term, we calculate its "instantiation frequency" as the percentage of abstracts containing the term in which at least one annotator assigned it to a span of text.
The total numbers of MeSH terms with an instantiation rate above different thresholds for the respective PIO elements are shown in Table~\ref{table:mesh_rates}.

\begin{table}
\small 
    \centering
    \begin{tabular}{ l c c c} 
    	\hline
        Inst. Freq & 10\% & 25\% & 50\% \\
        \hline
        Participants & 65 & 24 & 7 \\
        Interventions & 106 & 68 & 32 \\
        Outcomes & 118 & 108 & 75 \\
        \hline
    \end{tabular}
    \caption{The number of common MeSH terms (out of 135) that were assigned to a span of text in at least 10\%, 25\%, and 50\% of the possible documents. } 
    	\label{table:mesh_rates}
\end{table}

\section{Tasks \& Baselines}
\label{section:tasks-baselines}

We outline a few NLP tasks that are central to the aim of processing medical literature generally and to aiding practitioners of EBM specifically. First, we consider the task of identifying spans in abstracts that describe the respective PICO elements (Section \ref{section:tasks-spans}). This would, e.g., improve medical literature search and retrieval systems. Next, we outline the problem of extracting structured information from abstracts (Section \ref{section:tasks-extraction}). Such models would further aid search, and might eventually facilitate automated knowledge-base construction for the clinical trials literature. Furthermore, automatic extraction of structured data would enable automation of the manual evidence synthesis process \cite{marshall:2017:ACL}.

Finally, we consider the challenging task of identifying redundant mentions of the same PICO element (Section \ref{section:tasks-repetition}). This happens, e.g., when an intervention is mentioned by the authors repeatedly in an abstract, potentially with different terms. Achieving such disambiguation is important for systems aiming to induce structured representations of trials and their results, as this would require recognizing and normalizing the unique interventions and outcomes studied in a trial.  




For each of these tasks we present baseline models and corresponding results. Note that we have pre-defined train, development and test sets across PIO elements for this corpus, comprising 4300, 500 and 200 abstracts, respectively. The latter set is annotated by domain experts (i.e., persons with medical training). These splits will, of course, be distributed along with the dataset to facilitate model comparisons.


\subsection{Identifying P, I and O Spans}
\label{section:tasks-spans}

We consider two baseline models: a linear Conditional Random Field (CRF) ~\cite{lafferty2001conditional} and a Long Short-Term Memory (LSTM) neural tagging model, an LSTM-CRF \cite{lample2016neural,ma-hovy:2016:P16-1}. In both models, we treat tokens as being either Inside (I) or Outside (O) of spans.  

For the CRF, features include: indicators for the current, previous and next words; part of speech tags inferred using the Stanford CoreNLP tagger \cite{DBLP:conf/acl/ManningSBFBM14}; and character information, e.g., whether a token contains digits, uppercase letters, symbols and so on.%


\begin{table}
\small
\centering
\begin{tabular}{ l c c c c}
\hline
\bf{CRF} & Precision & Recall & F-1 \\
\cline{2-4}
Participants & 0.55 & 0.51 & 0.53 \\
Interventions & 0.65 & 0.21 & 0.32 \\
Outcomes & 0.83&0.17&0.29\\
\hline
\bf{LSTM-CRF} & Precision & Recall & F-1 \\
\cline{2-4}
Participants & 0.78&0.66&0.71 \\
Interventions & 0.61&0.70&0.65 \\
Outcomes & 0.69&0.58&0.63 \\
\hline
\end{tabular}
\caption{Baseline models (on the test set) for the PIO span tagging task.}
\end{table}
For the neural model, the model induces features via a bi-directional LSTM that consumes distributed vector representations of input tokens sequentially. The bi-LSTM yields a hidden vector at each token index, which is then passed to a CRF layer for prediction. We also exploit character-level information by passing a bi-LSTM over the characters comprising each word ~\cite{lample2016neural}; these are appended to the word embedding representations before being passed through the bi-LSTM. 





\subsection{Extracting Structured Information}
\label{section:tasks-extraction}

Beyond identifying the spans of text containing information pertinent to each of the PIO elements, we consider the task of predicting which of the detailed labels occur in each span, and where they are located.
Specifically, we begin with the starting spans and predict a single label from the corresponding PIO hierarchy for each token, evaluating against the test set of 200 documents.
Initial experiments with neural models proved unfruitful but bear further investigation.

For the CRF model we include the same features as in the previous model, supplemented with additional features encoding if the adjacent tokens include any parenthesis or mathematical operators (specifically: $\%, +, -$).
For the logistic regression model, we use a one-vs-rest approach.
Features include token $n$-grams, part of speech indicators, and the same character-level information as in the CRF model. 

\begin{table}
\small
    \centering
    \begin{tabular}{ l c c c} 
    	\hline
        \textbf{LogReg} & Precision & Recall & F-1 \\
        \cline{2-4}
        Participants & 0.41 & 0.20 & 0.26 \\
        Interventions & 0.79 & 0.44 & 0.57 \\
        Outcomes & 0.24 & 0.21 & 0.22 \\
        \hline
        \textbf{CRF} & Precision & Recall & F-1 \\
        \cline{2-4}
        Participants & 0.41 & 0.25 & 0.31 \\
        Interventions & 0.59 & 0.15 & 0.21 \\
        Outcomes & 0.60 & 0.51 & 0.55\\
       	\hline
    \end{tabular}
    \caption{Baseline models for the token-level, detailed labeling task.}
\end{table}

\subsection{Detecting Repetition}
\label{section:tasks-repetition}

To formalize repetition, we consider every pair of starting PIO spans from each abstract, and assign binary labels that indicate whether they share at least one instance of the same information.
Although this makes prediction easier for long and information-dense spans, a large enough majority of the spans contain only a single instance of relevant information that the task serves as a reasonable baseline.
Again, the model is trained on the aggregated labels collected from AMT and evaluated against the high-quality test set. 

We train a logistic regression model that operates over standard features, including bag-of-words representations and sentence-level features such as length and position in the document. All baseline model implementations are available on the corpus website. 

\begin{table}
\small
    \centering
    \begin{tabular}{ l c c c} 
        \textbf{} & Precision & Recall & F-1 \\
        \cline{2-4}
        \noalign{\vskip 1mm}  
        Participants & 0.39 & 0.52 & 0.44 \\
        Interventions & 0.41 & 0.50 & 0.45 \\
        Outcomes & 0.10 & 0.16 & 0.12 \\
	\end{tabular}
    \caption{Baseline model for predicting whether pairs of spans contain redundant information.}
\end{table}

\section{Conclusions}
\label{section:conclusions}

We have presented EBM-NLP: a new, publicly available corpus comprising 5,000 richly annotated abstracts of articles describing clinical randomized controlled trials. This dataset fills a need for larger scale corpora to facilitate research on NLP methods for processing the biomedical literature, which have the potential to aid the conduct of EBM. The need for such technologies will only become more pressing as the literature continues its torrential growth.  

The EBM-NLP corpus, accompanying documentation, code for working with the data, and baseline models presented in this work are all publicly available at: \url{http://www.ccs.neu.edu/home/bennye/EBM-NLP}. 

\section{Acknowledgements}

This work was supported in part by the National
Cancer Institute (NCI) of the National Institutes of
Health (NIH), award number UH2CA203711. 
\bibliography{acl2018}

\begin{thebibliography}{37}
\expandafter\ifx\csname natexlab\endcsname\relax\def\natexlab#1{#1}\fi

\bibitem[{Abacha and Zweigenbaum(2015)}]{abacha2015means}
Asma~Ben Abacha and Pierre Zweigenbaum. 2015.
\newblock Means: A medical question-answering system combining nlp techniques
  and semantic web technologies.
\newblock \emph{Information processing \& management}, 51(5):570--594.

\bibitem[{Alamri and Stevenson(2015)}]{alamri2015automatic}
Abdulaziz Alamri and Mark Stevenson. 2015.
\newblock Automatic detection of answers to research questions from medline.
\newblock \emph{Proceedings of the workshop on Biomedical Natural Language
  Processing (BioNLP)}, pages 141--146.

\bibitem[{Alamri and Stevenson(2016)}]{alamri2016corpus}
Abdulaziz Alamri and Mark Stevenson. 2016.
\newblock A corpus of potentially contradictory research claims from
  cardiovascular research abstracts.
\newblock \emph{Journal of biomedical semantics}, 7(1):36.

\bibitem[{Bastian et~al.(2010)Bastian, Glasziou, and
  Chalmers}]{bastian2010seventy}
Hilda Bastian, Paul Glasziou, and Iain Chalmers. 2010.
\newblock Seventy-five trials and eleven systematic reviews a day: how will we
  ever keep up?
\newblock \emph{PLoS medicine}, 7(9):e1000326.

\bibitem[{Borah et~al.(2017)Borah, Brown, Capers, and
  Kaiser}]{borah2017analysis}
Rohit Borah, Andrew~W Brown, Patrice~L Capers, and Kathryn~A Kaiser. 2017.
\newblock Analysis of the time and workers needed to conduct systematic reviews
  of medical interventions using data from the prospero registry.
\newblock \emph{BMJ open}, 7(2):e012545.

\bibitem[{Boudin et~al.(2010)Boudin, Nie, and Dawes}]{boudin2010positional}
Florian Boudin, Jian-Yun Nie, and Martin Dawes. 2010.
\newblock Positional language models for clinical information retrieval.
\newblock In \emph{Proceedings of the 2010 Conference on Empirical Methods in
  Natural Language Processing}, pages 108--115. Association for Computational
  Linguistics.

\bibitem[{Chung(2009)}]{chung2009sentence}
Grace~Y Chung. 2009.
\newblock Sentence retrieval for abstracts of randomized controlled trials.
\newblock \emph{{BMC} medical informatics and decision making}, 9(1):10.

\bibitem[{Dalvi et~al.(2013)Dalvi, Dasgupta, Kumar, and
  Rastogi}]{dalvi2013aggregating}
Nilesh Dalvi, Anirban Dasgupta, Ravi Kumar, and Vibhor Rastogi. 2013.
\newblock Aggregating crowdsourced binary ratings.
\newblock In \emph{Proceedings of the International Conference on World Wide
  Web (WWW)}, pages 285--294. ACM.

\bibitem[{Dawid and Skene(1979)}]{dawid1979maximum}
Alexander~Philip Dawid and Allan~M Skene. 1979.
\newblock Maximum likelihood estimation of observer error-rates using the em
  algorithm.
\newblock \emph{Applied statistics}, pages 20--28.

\bibitem[{Demner-Fushman and Lin(2007)}]{demner2007answering}
Dina Demner-Fushman and Jimmy Lin. 2007.
\newblock Answering clinical questions with knowledge-based and statistical
  techniques.
\newblock \emph{Computational Linguistics}, 33(1):63--103.

\bibitem[{Ferracane et~al.(2016)Ferracane, Marshall, Wallace, and
  Erk}]{ferracane2016leveraging}
Elisa Ferracane, Iain Marshall, Byron~C Wallace, and Katrin Erk. 2016.
\newblock Leveraging coreference to identify arms in medical abstracts: An
  experimental study.
\newblock In \emph{Proceedings of the Seventh International Workshop on Health
  Text Mining and Information Analysis}, pages 86--95.

\bibitem[{Fraser and Dunstan(2010)}]{fraser2010impossibility}
Alan~G Fraser and Frank~D Dunstan. 2010.
\newblock On the impossibility of being expert.
\newblock \emph{British Medical Journal}, 341:c6815.

\bibitem[{Hovy et~al.(2014)Hovy, Plank, and S{\o}gaard}]{hovy2014experiments}
Dirk Hovy, Barbara Plank, and Anders S{\o}gaard. 2014.
\newblock Experiments with crowdsourced re-annotation of a pos tagging data
  set.
\newblock In \emph{Proceedings of the 52nd Annual Meeting of the Association
  for Computational Linguistics (volume 2: Short Papers)}, volume~2, pages
  377--382.

\bibitem[{Huang et~al.(2006)Huang, Lin, and
  Demner-Fushman}]{huang2006evaluation}
Xiaoli Huang, Jimmy Lin, and Dina Demner-Fushman. 2006.
\newblock Evaluation of {PICO} as a knowledge representation for clinical
  questions.
\newblock In \emph{AMIA annual symposium proceedings}, volume 2006, page 359.
  American Medical Informatics Association.

\bibitem[{Jonnalagadda et~al.(2015)Jonnalagadda, Goyal, and
  Huffman}]{jonnalagadda2015automating}
Siddhartha~R Jonnalagadda, Pawan Goyal, and Mark~D Huffman. 2015.
\newblock Automating data extraction in systematic reviews: a systematic
  review.
\newblock \emph{Systematic reviews}, 4(1):78.

\bibitem[{Kiritchenko et~al.(2010)Kiritchenko, de~Bruijn, Carini, Martin, and
  Sim}]{kiritchenko2010exact}
Svetlana Kiritchenko, Berry de~Bruijn, Simona Carini, Joel Martin, and Ida Sim.
  2010.
\newblock Exact: automatic extraction of clinical trial characteristics from
  journal publications.
\newblock \emph{BMC medical informatics and decision making}, 10(1):56.

\bibitem[{Lafferty et~al.(2001)Lafferty, McCallum, and
  Pereira}]{lafferty2001conditional}
John Lafferty, Andrew McCallum, and Fernando~CN Pereira. 2001.
\newblock Conditional random fields: Probabilistic models for segmenting and
  labeling sequence data.

\bibitem[{Lample et~al.(2016)Lample, Ballesteros, Subramanian, Kawakami, and
  Dyer}]{lample2016neural}
Guillaume Lample, Miguel Ballesteros, Sandeep Subramanian, Kazuya Kawakami, and
  Chris Dyer. 2016.
\newblock Neural architectures for named entity recognition.
\newblock In \emph{Proceedings of NAACL-HLT}, pages 260--270.

\bibitem[{Lowe and Barnett(1994)}]{lowe1994understanding}
Henry~J Lowe and G~Octo Barnett. 1994.
\newblock Understanding and using the medical subject headings (mesh)
  vocabulary to perform literature searches.
\newblock \emph{Jama}, 271(14):1103--1108.

\bibitem[{Lu et~al.(2009)Lu, Kim, and Wilbur}]{lu2009evaluation}
Zhiyong Lu, Won Kim, and W~John Wilbur. 2009.
\newblock Evaluation of query expansion using mesh in pubmed.
\newblock \emph{Information retrieval}, 12(1):69--80.

\bibitem[{Ma and Hovy(2016)}]{ma-hovy:2016:P16-1}
Xuezhe Ma and Eduard Hovy. 2016.
\newblock \href {http://www.aclweb.org/anthology/P16-1101} {End-to-end sequence
  labeling via bi-directional lstm-cnns-crf}.
\newblock In \emph{Proceedings of the 54th Annual Meeting of the Association
  for Computational Linguistics (Volume 1: Long Papers)}, pages 1064--1074,
  Berlin, Germany. Association for Computational Linguistics.

\bibitem[{Manning et~al.(2014)Manning, Surdeanu, Bauer, Finkel, Bethard, and
  McClosky}]{DBLP:conf/acl/ManningSBFBM14}
Christopher~D. Manning, Mihai Surdeanu, John Bauer, Jenny~Rose Finkel, Steven
  Bethard, and David McClosky. 2014.
\newblock The stanford corenlp natural language processing toolkit.
\newblock In \emph{Proceedings of the 52nd Annual Meeting of the Association
  for Computational Linguistics, {ACL} 2014, June 22-27, 2014, Baltimore, MD,
  USA, System Demonstrations}, pages 55--60.

\bibitem[{Marshall et~al.(2017)Marshall, Kuiper, Banner, and
  Wallace}]{marshall:2017:ACL}
Iain Marshall, Jo{\"e}l Kuiper, Edward Banner, and Byron~C. Wallace. 2017.
\newblock \href {https://doi.org/10.18653/v1/P17-4002} {{Automating Biomedical
  Evidence Synthesis: RobotReviewer}}.
\newblock In \emph{Proceedings of the Association for Computational Linguistics
  (ACL), System Demonstrations}, pages 7--12. Association for Computational
  Linguistics (ACL).

\bibitem[{Moll{\'a} and Santiago-Martinez(2011)}]{molla2011development}
Diego Moll{\'a} and Maria~Elena Santiago-Martinez. 2011.
\newblock Development of a corpus for evidence based medicine summarisation.

\bibitem[{Mortensen et~al.(2017)Mortensen, Adam, Trikalinos, Kraska, and
  Wallace}]{mortensen2017exploration}
Michael~L Mortensen, Gaelen~P Adam, Thomas~A Trikalinos, Tim Kraska, and
  Byron~C Wallace. 2017.
\newblock An exploration of crowdsourcing citation screening for systematic
  reviews.
\newblock \emph{Research synthesis methods}, 8(3):366--386.

\bibitem[{Nguyen et~al.(2017)Nguyen, Wallace, Li, Nenkova, and
  Lease}]{nguyen2017aggregating}
An~T Nguyen, Byron~C Wallace, Junyi~Jessy Li, Ani Nenkova, and Matthew Lease.
  2017.
\newblock Aggregating and predicting sequence labels from crowd annotations.
\newblock In \emph{Proceedings of the conference. Association for Computational
  Linguistics. Meeting}, volume 2017, page 299. NIH Public Access.

\bibitem[{Novotney and Callison-Burch(2010)}]{novotney2010cheap}
Scott Novotney and Chris Callison-Burch. 2010.
\newblock Cheap, fast and good enough: Automatic speech recognition with
  non-expert transcription.
\newblock In \emph{Human Language Technologies: The 2010 Annual Conference of
  the North American Chapter of the Association for Computational Linguistics},
  pages 207--215. Association for Computational Linguistics.

\bibitem[{Sabou et~al.(2012)Sabou, Bontcheva, and
  Scharl}]{sabou2012crowdsourcing}
Marta Sabou, Kalina Bontcheva, and Arno Scharl. 2012.
\newblock Crowdsourcing research opportunities: lessons from natural language
  processing.
\newblock In \emph{Proceedings of the 12th International Conference on
  Knowledge Management and Knowledge Technologies}, page~17. ACM.

\bibitem[{Scells et~al.(2017)Scells, Zuccon, Koopman, Deacon, Azzopardi, and
  Geva}]{scells2017test}
Harrisen Scells, Guido Zuccon, Bevan Koopman, Anthony Deacon, Leif Azzopardi,
  and Shlomo Geva. 2017.
\newblock A test collection for evaluating retrieval of studies for inclusion
  in systematic reviews.
\newblock In \emph{Proceedings of the 40th International ACM SIGIR Conference
  on Research and Development in Information Retrieval}, pages 1237--1240. ACM.

\bibitem[{Stenetorp et~al.(2012)Stenetorp, Pyysalo, Topi{\'c}, Ohta, Ananiadou,
  and Tsujii}]{stenetorp2012brat}
Pontus Stenetorp, Sampo Pyysalo, Goran Topi{\'c}, Tomoko Ohta, Sophia
  Ananiadou, and Jun'ichi Tsujii. 2012.
\newblock Brat: a web-based tool for nlp-assisted text annotation.
\newblock In \emph{Proceedings of the Demonstrations at the 13th Conference of
  the European Chapter of the Association for Computational Linguistics}, pages
  102--107. Association for Computational Linguistics.

\bibitem[{Summerscales et~al.(2011)Summerscales, Argamon, Bai, Hupert, and
  Schwartz}]{summerscales2011automatic}
Rodney~L Summerscales, Shlomo Argamon, Shangda Bai, Jordan Hupert, and Alan
  Schwartz. 2011.
\newblock Automatic summarization of results from clinical trials.
\newblock In \emph{Bioinformatics and Biomedicine (BIBM), 2011 IEEE
  International Conference on}, pages 372--377. IEEE.

\bibitem[{Thomas et~al.(2017)Thomas, Noel-Storr, Marshall, Wallace, McDonald,
  Mavergames, Glasziou, Shemilt, Synnot, Turner et~al.}]{thomas2017living}
James Thomas, Anna Noel-Storr, Iain Marshall, Byron Wallace, Steven McDonald,
  Chris Mavergames, Paul Glasziou, Ian Shemilt, Anneliese Synnot, Tari Turner,
  et~al. 2017.
\newblock Living systematic reviews: 2. combining human and machine effort.
\newblock \emph{Journal of clinical epidemiology}, 91:31--37.

\bibitem[{Tsafnat et~al.(2013)Tsafnat, Dunn, Glasziou, Coiera
  et~al.}]{tsafnat2013automation}
Guy Tsafnat, Adam Dunn, Paul Glasziou, Enrico Coiera, et~al. 2013.
\newblock The automation of systematic reviews.
\newblock \emph{BMJ}, 346(f139):1--2.

\bibitem[{Verbeke et~al.(2012)Verbeke, Van~Asch, Morante, Frasconi, Daelemans,
  and De~Raedt}]{verbeke2012statistical}
Mathias Verbeke, Vincent Van~Asch, Roser Morante, Paolo Frasconi, Walter
  Daelemans, and Luc De~Raedt. 2012.
\newblock A statistical relational learning approach to identifying evidence
  based medicine categories.
\newblock In \emph{Proceedings of the Joint Conference on Empirical Methods in
  Natural Language Processing and Computational Natural Language Learning},
  pages 579--589. Association for Computational Linguistics.

\bibitem[{Wallace et~al.(2013)Wallace, Dahabreh, Schmid, Lau, and
  Trikalinos}]{wallace2013modernizing}
Byron~C Wallace, Issa~J Dahabreh, Christopher~H Schmid, Joseph Lau, and
  Thomas~A Trikalinos. 2013.
\newblock Modernizing the systematic review process to inform comparative
  effectiveness: tools and methods.
\newblock \emph{Journal of comparative effectiveness research}, 2(3):273--282.

\bibitem[{Wallace et~al.(2016)Wallace, Kuiper, Sharma, Zhu, and
  Marshall}]{wallace2016extracting}
Byron~C Wallace, Jo{\"e}l Kuiper, Aakash Sharma, Mingxi~Brian Zhu, and Iain~J
  Marshall. 2016.
\newblock Extracting {PICO} sentences from clinical trial reports using
  supervised distant supervision.
\newblock \emph{Journal of Machine Learning Research}, 17(132):1--25.

\bibitem[{Wallace et~al.(2017)Wallace, Noel-Storr, Marshall, Cohen, Smalheiser,
  and Thomas}]{wallace2017identifying}
Byron~C Wallace, Anna Noel-Storr, Iain~J Marshall, Aaron~M Cohen, Neil~R
  Smalheiser, and James Thomas. 2017.
\newblock Identifying reports of randomized controlled trials (rcts) via a
  hybrid machine learning and crowdsourcing approach.
\newblock \emph{Journal of the American Medical Informatics Association},
  24(6):1165--1168.

\end{thebibliography}
\bibliographystyle{acl_natbib}

\end{document}